\documentclass[a4paper]{svproc}
\usepackage{graphics} 
\usepackage{epsfig} 
\usepackage{mathptmx} 
\usepackage{times} 
\usepackage{amsmath} 
\usepackage{amssymb}  
\usepackage{multirow}

\usepackage{bm}

\usepackage[english]{babel}
\usepackage{cite}
\usepackage{subfigure}

\usepackage{todonotes}

\usepackage{url}

\usepackage{times}
\usepackage{amsfonts}
\usepackage{mathtools}
\usepackage{color}
\usepackage{xspace} 
\usepackage[bookmarks=true]{hyperref}
\usepackage{setspace}

\usepackage{tikz,pgf}
\usetikzlibrary{arrows,automata,shapes,calc,backgrounds,spy,positioning}

\usepackage{epstopdf}

\usepackage{cite}
\newcommand{\etal}{\textit{et al}. }

\makeatletter
\newcommand{\printfnsymbol}[1]{%
  \textsuperscript{\@fnsymbol{#1}}%
}
\makeatother

\graphicspath{ {figs/} {figs/ssp_raptor_s_plots/} {figs/nocs/} {figs/quad_person_occlusion/} {figs/headline_figs/} {figs/overview_figs/} {figs/ssp_fail/} }  

\usepackage{url}

\begin{document}
\mainmatter              
\title{MSL-RAPTOR: A 6DoF Relative Pose Tracker for Onboard Robotic Perception}
\titlerunning{MSL-RAPTOR}  
%
\author{Benjamin Ramtoula\thanks{Equal contribution}\inst{1} \and Adam Caccavale\printfnsymbol{1}\inst{2} \and Giovanni Beltrame\inst{3} \and Mac Schwager\inst{2}}
\authorrunning{B. Ramtoula et al.} 
%
\tocauthor{Benjamin Ramtoula, Adam Caccavale, Giovanni Beltrame, and Mac Schwager}
\institute{
		University of Oxford, Oxford, UK\\
		\email{benjamin@robots.ox.ac.uk}
	\and
        Stanford University, Stanford, USA\\
        \email{\{awc11,schwager\}@stanford.edu}
    \and
        Polytechnique Montréal, Montréal, Canada\\
		\email{giovanni.beltrame@polymtl.ca}}

\maketitle              

\begin{abstract}
  Determining the relative position and orientation of objects in an
  environment is a fundamental building block for a wide range of
  robotics applications. To accomplish this task efficiently in
  practical settings, a method must be fast, use common sensors, and
  generalize easily to new objects and environments. We present
  MSL-RAPTOR, a two-stage algorithm for tracking a rigid body with a
  monocular camera. The image is first processed by an efficient
  neural network-based front-end to detect new objects and track 2D
  bounding boxes between frames. The class label and bounding box is
  passed to the back-end that updates the object's pose using an
  unscented Kalman filter (UKF). The measurement posterior is fed back
  to the 2D tracker to improve robustness. The object's class is
  identified so a class-specific UKF can be used if custom dynamics
  and constraints are known. Adapting to track the pose of new classes
  only requires providing a trained 2D object detector or labeled 2D
  bounding box data, as well as the approximate size of the
  objects. The performance of MSL-RAPTOR is first verified on the
  NOCS-REAL275 dataset, achieving results comparable to RGB-D
  approaches despite not using depth measurements. When tracking a
  flying drone from onboard another drone, it outperforms the fastest
  comparable method in speed by a factor of 3, while giving lower
  translation and rotation median errors by 66\% and 23\%
  respectively.  
\end{abstract}

\section{Introduction}



\begin{figure}[t]
\centering
\includegraphics[clip, trim=0cm 2.75cm 0cm 0cm, width=.95\linewidth]{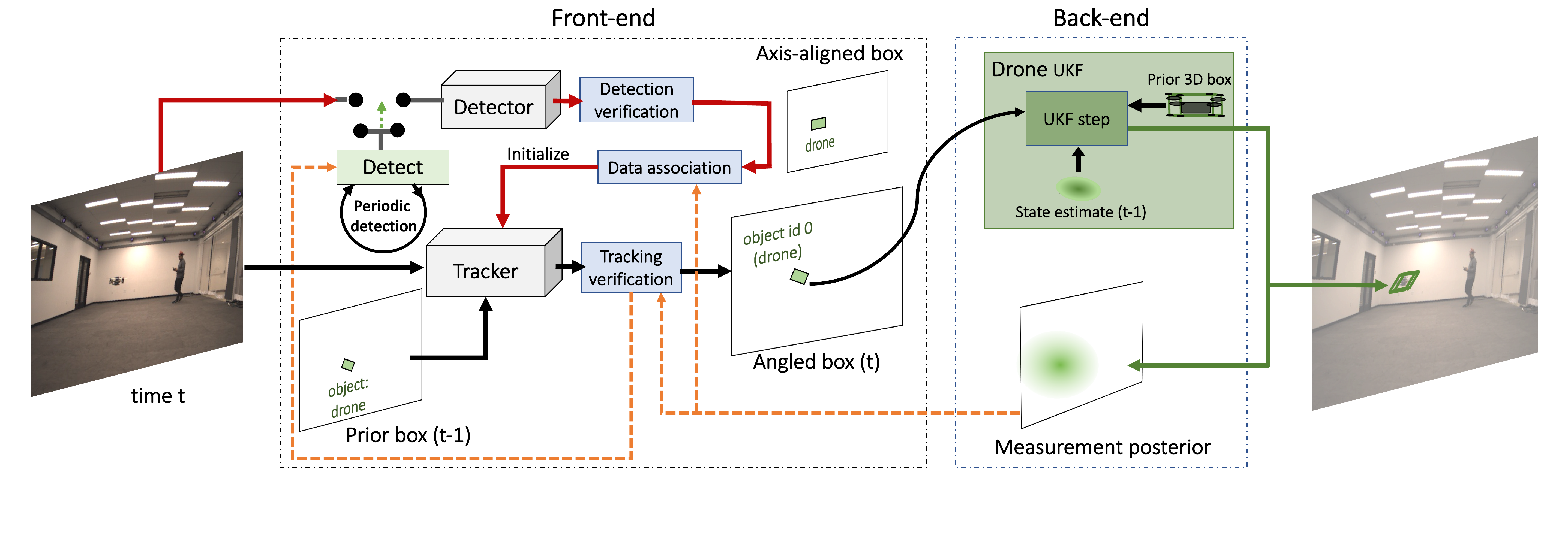}
\caption{Overview of MSL-RAPTOR. The front-end generates angled-bounding boxes \& class labels from monocular images. The back-end tracks the object's pose with a class-specific UKF, and passes its uncertainty metrics to the front-end to detect when the tracker has lost the object.}
\label{fig:overview}
\end{figure}


For a robot to intelligently interact with its environment, it must
quickly and accurately determine the relative position and orientation
of neighboring robots, people, vehicles, and a variety of other static
and dynamic obstacles in the scene. From drone
racing~\cite{spica2018realtime} to pedestrian
avoidance~\cite{badue2019selfdrivingcarsurvey}, tasks relying on such
estimates are numerous and are essential to creating truly autonomous
robots. Regardless, the ability to track the six Degree of Freedom
(6DoF) relative pose of a dynamic object in real-time, with onboard
sensing and computation, is largely beyond the reach of existing robot
perception frameworks.  To fill this need, we propose
\textbf{M}onocular \textbf{S}equential \textbf{L}ightweight
\textbf{R}otation \textbf{A}nd \textbf{P}osition \textbf{T}racking
\textbf{O}n-\textbf{R}obots (MSL-RAPTOR), a lightweight,
general-purpose robot perception module for tracking the 6DoF relative
pose of a dynamic object using only an onboard monocular
camera. MSL-RAPTOR performs similarly to state-of-the-art approaches
relying on more sensors (e.g. depth) and greater computation
capability.

Existing methods for 6DoF relative pose estimation from vision can be
divided into model-based, deep learning-based, and hybrid methods.
Existing model-based approaches rely on matching the view of an object
to a database of templates or visual features. These methods are
limited in the face of scenes with low textures or changing
conditions, and cannot be used when prior 3D models of objects are not
available \cite{schmidt2014dart,andriluka2010humantracking}. Deep
learning methods such as PoseCNN~\cite{xiang2017posecnn} and
LieNets~\cite{do2018lienets} overcome these challenges by directly
estimating the relative pose from images, learning this inverse map
from a large corpus of training data.  Hybrid approaches combine
elements of deep neural networks and model-based methods.  For
example, BB8~\cite{rad2017bb8}, SSD-6D~\cite{kehl2017ssd}, and
SingleShotPose~\cite{tekin2018ssp} (SSP) all use deep neural networks
to estimate the 2D projections of the 3D bounding box encompassing the
object.  These projections are then used to solve the
Perspective-n-Point (PnP) problem to obtain a pose
estimate~\cite{lepetit2009epnp}. Another hybrid method,
NOCS~\cite{wang2019normalized}, estimates pose using pixel
correspondences and depth data with category-level canonical
representations.  However, the above deep and hybrid methods ignore
temporal correlations among multiple frames to aid in tracking
objects.  In contrast, 6-PACK~\cite{wang20196sixpack} leverages
temporal consistency by learning to represent objects using keypoints
that are matched between frames. While these methods present
impressive pose estimation performance, they are either too slow, too
computationally demanding, or require too much prior model knowledge
to be used for practical onboard relative pose perception for
resource-limited robots.

These same limitations apply to the many pose tracking approaches
developed for self-driving cars, for which tracking methods represent
a key component of the
system~\cite{badue2019selfdrivingcarsurvey}. These rely on the use of
rich sensors and/or advanced computational capabilities to achieve
online performance and often assume planar constraints since they deal
solely with land-based objects (e.g. pedestrians, other vehicles),
hence are only SE(2) trackers with 3DoF (e.g.~\cite{cho2014multi,
  buyval2018realtime}).

\textbf{Contribution.} With the requirements for onboard robotic
perception in mind, we propose MSL-RAPTOR, which seeks to combine the
best aspects of deep learning-based and model-based perception
paradigms.  MSL-RAPTOR takes raw image data with a dual deep neural
network architecture on the front-end to produce and track an angled
2D bounding box around an object in the scene.  A single 2D bounding
box is not enough information to determine the object's pose (see
Fig.~\ref{fig:pose_ambiguity-a}), so this semantically labeled
bounding box is passed to a model-based Unscented Kalman Filter
(UKF)~\cite{julier2000ukf, wan2000ukf} back-end to fuse measurements
across time and produce a Bayesian relative pose estimate for the
object.  The deep front-end uses a YOLOv3
network~\cite{redmon2018yolov3}, retrained with robot-specific object
classes, for initial object detection and labeling, together with a
SiamMask network~\cite{wang2019siammask} for quickly tracking the 2D
object bounding box from frame to frame once it has been detected.
The 2D bounding box corners and semantic object labels (e.g.,
``person", ``UAV", ``car", ``tree") are then passed to the UKF
back-end.  The back-end uses the bounding box corners as
pseudo-measurements, which are fused over time to infer the full 6DoF
relative pose of the object.  The UKF propagates the target's motion
with a label-specific motion model (since people, cars, and UAVs all
move differently).  We then close the loop between back-end and
front-end with a re-detection trigger.  If the Mahalanobis distance
between the measurement and its posterior is above a threshold or if
the pose estimate is too close to the edge of the field of view
(judged using the z-test based on its covariance), the target is
considered lost, and we re-trigger the YOLO target detection on the
front-end.  With this framework, we can track an object through
occlusions and temporary absences from the field of view.  MSL-RAPTOR
currently assumes a single object in the scene, but the extension to
multiple objects (and classes) is straightforward.

We demonstrate MSL-RAPTOR running on a Jetson TX2 onboard a standard
quadrotor UAV platform while tracking another UAV at 8.7 frames per
second. This is 3 times faster than the fastest comparable state-of-the-art
method, SSP~\cite{tekin2018ssp}, and produces pose estimates with 66\%
and 23\% lower translation and rotation median errors, respectively.
We also compare performance with several state-of-the-art deep
learning methods for 6DoF pose tracking using a depth camera.  We show
on standard datasets that MSL-RAPTOR performs similarly to these
existing methods despite only using a monocular camera with no depth
information.



\section{Problem Formulation}


We consider the scenario where a computationally-limited mobile
robotic agent with a calibrated monocular camera needs to identify and
track the relative pose of a dynamic object $O$ from a series of RGB
images $I^t \in \mathbb{R}^{W \times H}$. The state at time $t$ is
$\mathbf{x}^t = [\mathbf{p}^t, \mathbf{v}^t, \mathbf{q}^t,
\bm{\omega}^t] \in\mathbb{R}^{13}$ where
$\mathbf{p}^t \in \mathbb{R}^3$ is 3D position,
$\mathbf{v}^t \in \mathbb{R}^3$ is linear velocity,
$\mathbf{q}^t \in \mathbb{R}\times\mathbb{R}^3$ is the orientation as
a quaternion, and $\bm{\omega}^t \in \mathbb{R}^3$ is angular
velocity. The position and orientation component of this state can be
represented as the object pose $\mathbf{T}^t \in SE(3)$. Henceforth
the time superscripts will be dropped for notational convenience
unless they are needed for clarity.

The object $O$ has a span in 3D space that can be approximated with a
sparse set of $n$ points $\mathbf{B} \in \mathbb{R}^{n\times3}$, most
commonly the corners of its 3D bounding box which we use by
default. Additionally, each object has a corresponding class label
(e.g. \texttt{drone}, \texttt{bottle}, etc) denoted $c$. Each class of
object has an associated dynamics model used to propagate the
components of the state in the UKF back-end:
$\mathbf{x}^{t_2} = \mathbf{F}_c(\mathbf{x}^{t_1})$. By default a
simple constant velocity model is used, but more sophisticated
class-specific models may be provided. In general we assume no
knowledge of the tracked object's control input.

\begin{figure}[!tpb]%
\centering
\subfigure[][]{%
    \centering
    \includegraphics[clip, trim = 0.25cm 0.1cm 0cm 0.1cm,width=0.6\linewidth]{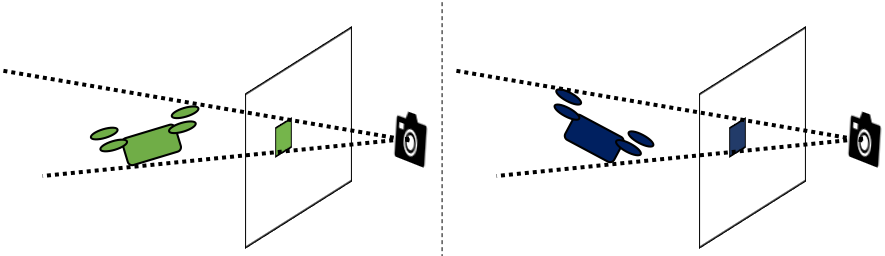}%
    \label{fig:pose_ambiguity-a} }%
\hspace{30pt}%
\subfigure[][]{%
    \centering
    \includegraphics[clip, trim = 14.75cm 7cm 2.5cm 3.5cm, width=0.15\linewidth]{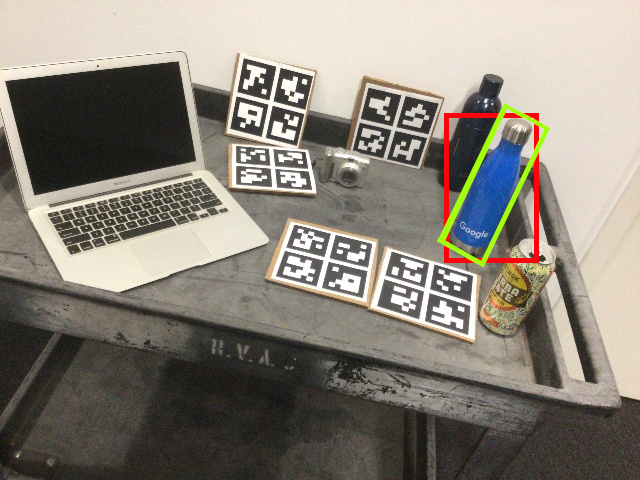} %
    \label{fig:bounding_box_types-b} }%
  \caption{Illustration of the pose ambiguity from bounding box
    measurements: \subref{fig:pose_ambiguity-a} same 2D bounding box
    produced from difference
    poses; 
    \subref{fig:bounding_box_types-b} pose ambiguity can be reduced
    with an angled bounding box, but not
    eliminated.}
\label{fig:obs_ambiguity}%
\end{figure}

When processing an image, the 2D projections of the object's points
$\mathbf{B}$ onto the image $I$ are enclosed in the minimum-area
\textit{angled} bounding box: $\mathbf{z} = [x, y, w, h, \alpha]$
where $x$ and $y$ are the column and row of the box center, $w$ and
$h$ are the box width and height, and $\alpha$ is the box angle from
the horizontal image axis. If instead an \textit{axis-aligned}
bounding box is produced (i.e. $\alpha=0$), it is denoted
$\mathbf{\tilde{z}}$ (Fig.~\ref{fig:bounding_box_types-b}). The
distribution of estimates over states and measurements are both
modeled as Gaussians, $\mathcal{N}(\mathbf{\hat{x}},\hat{\Sigma})$ and
$\mathcal{N}(\mathbf{\hat{z}},\mathbf{\hat{S}})$ respectively. When
evaluating the consistency of a measurement $\mathbf{z}^t$ with our
prior estimate, we use the Mahalanobis distance defined as:
\begin{equation}
    \label{eqn:mahalanobis}
    \mathbb{M}(\mathbf{z^t}, \mathbf{\hat{z}^{t-1}}, \mathbf{\hat{S}^{t-1}}) \doteq \sqrt{\left(\mathbf{z^{t}}-\hat{\mathbf{z}}^{t-1}\right)^T \left(\mathbf{\hat{S}^{t-1}}\right)^{-1} \left(\mathbf{z}^{t}-\hat{\mathbf{z}}^{t-1}\right)}.
\end{equation}

\section{Algorithm}


The architecture of our approach is illustrated in
Fig.~\ref{fig:overview}. The front-end has two components for
processing the images: a detector and a visual tracker. The detector
can identify all objects in an image and their class labels, but is
slower and only generates axis-aligned bounding boxes. The tracker is
faster and produces angled bounding boxes, but requires an accurate 2D
bounding box from the previous timestep. The back-end relies on a
class-specific UKF, taking in an angled bounding box measurement from
the front-end to produce pose estimates. For each image, the decision
to use the tracker vs. detection is determined by the consistency of
the new bounding box measurement with the measurement posterior
generated by the back-end's UKF.

\subsection{Object detection}
\label{subsec:new_detection}
Upon receiving an image $I$, the angled bounding boxes must be
extracted. If the object has not been previously located, or if
re-detection has been triggered, a pre-trained, out-of-the-box object
detector outputs the axis-aligned bounding box $\tilde{\mathbf{z}}$
and class label $c$ for each visible object $O_i$:
$\mathbb{D}(I)~\rightarrow~\{\mathbf{\tilde{z}}_{i}, c_{i} \}$.

Only bounding boxes for the class of interest are considered, and if
multiple boxes are found, the one closest to our current estimate as
measured by Eqn.~\ref{eqn:mahalanobis} is used. If our estimate's
uncertainty is too high, then the box with highest confidence is
chosen. This axis-aligned bounding box is then used to initialize the
visual tracker (Sec.~\ref{subsec:visual_tracking}) which produces an
angled bounding box measurement $\mathbf{z}$. This, along with the
class label $c$ from the detector, serves to initialize the UKF
(Sec.~\ref{subsec:ukf}). Because the filter requires an initial pose
estimate, the initial 6D pose is roughly approximated assuming the
width, height, and roll are aligned with the 2D bounding box. This is
a poor assumption in general, but the robustness of the filter will
compensate for the initialization error. As the class of the object is
also identified, class-based initialization rules can also be used
(e.g. mugs are usually upright).

\subsection{Visual tracking}
\label{subsec:visual_tracking}
Once an object has been detected and the associated tracker and UKF
are initialized, MSL-RAPTOR's front-end only relies on the tracker to
provide measurements to the UKF.  The tracker is an object-agnostic
method, which produces an angled bounding box measurement given the
previous bounding box and a new image
$\mathbb{T}(I^t, \mathbf{z}^{t-1}) \rightarrow
\mathbf{z}^t$. 

When a new image is available, the tracker generates an angled
bounding box that serves as an observation for the UKF predicting the
object's pose (Sec.~\ref{subsec:ukf}).

Aside from the speed benefit, we use a visual tracking method that
produces angled bounding boxes that are more informative measurements
than the axis-aligned boxes as illustrated in
Fig.~\ref{fig:obs_ambiguity}. While the angle is still not sufficient
to uniquely determine an object's pose, it reduces the set of poses
that would result in the given measurement which improves tracking
performance. 

\subsection{Unscented Kalman Filter}
\label{subsec:ukf}
To estimate pose from a series of angled bounding box measurements, we
use a class-dependent unscented Kalman filter. Probabilistic filters
like the UKF consist of a prediction and update step. During
prediction, the state is propagated forward using the object dynamics,
and the expected measurement from the resulting anticipated state is
calculated. In the update step, the difference between the predicted
and actual measurement is used to adjust the state estimate and
uncertainty. A typical output of a UKF is the mean and covariance of
the state, but as part of the update process, the measurement mean and
covariance are also calculated. These are passed to the front-end for
determining whether to trigger re-detection
(Sec.~\ref{subsec:update_tracked_objects}).

The prediction step of the UKF relies on a dynamics model, which can
optionally depend on the object tracked. The MSL-RAPTOR framework
gives access to a class label for each object, enabling the use of
class-specific dynamics if available, or a simple constant velocity
model if not. For systems without significant inertial effects this is
a good approximation, otherwise, we rely on the robustness of the
filter to correct for the modeling error.

To predict the expected measurement associated to a pose, the
knowledge of the 3D points representing the tracked object is
used. The vertices $\mathbf{B}$ are projected into the image plane
using the known camera intrinsic matrix $\mathbf{K}$ and the estimate
of the object's pose in the camera frame $\mathbf{T}$
\cite{Hartley2004compviz}. The minimum area rectangle that encloses
these projected pixels is the predicted measurement
$\mathbb{\pi}_\mathbf{K}(\mathbf{B}, \mathbf{T}) \rightarrow
\mathbf{z}$.

A UKF is chosen over alternatives such as an EKF or PF due to its
efficient handling of the non-linearities inherent in the system. The
PF's natural ability to track multiple-hypothesis potentially could
allow it to better determine poses that best explain the series of
bounding box measurements, but empirical testing showed the
computational requirements of this method to be too high to run
onboard.

\subsection{Updating the tracked objects}
\label{subsec:update_tracked_objects}
Occlusions, objects leaving the field of view, and poor measurements
(e.g. due to light reflections) can all cause tracking failures that
lead to inaccurate bounding boxes. No assumptions are made regarding
these scenarios. To address these challenges, detection is triggered
periodically to identify and locate when the object has become
unoccluded or has entered the image. This has the advantage of
compensating for any drift or errors that might occur in the visual
tracking. The front-end's verification mechanisms can also trigger
detection to correct for uncertain measurements.



\subsubsection{Bounding Box Verification}
\paragraph{Detected measurements}
Detection returns both the bounding box and the class label of the
object. This can be used to filter detections by class and to set
class-specific rejection conditions for the measurement that can help
prevent false-positive matches. For example, an aspect ratio condition
can be set so that a bounding box $\mathbf{\tilde{z}}$ is valid only
if it is in a valid class-specific aspect-ratio range:
\begin{equation*}
    \Gamma_\text{min}(c) < \tilde{w} / \tilde{h} < \Gamma_\text{max}(c), 
\end{equation*}
where the thresholds $\Gamma_\text{min}(c)$ and $\Gamma_\text{max}(c)$ can be determined from the typical geometry of the class. Additionally, a threshold on the minimum distance $d_\text{min}$ to the edge of the image can prevent bad measurements due to truncated bounding boxes:
\begin{align*}
    d_\text{min} < \tilde{x} - \tilde{w} / 2 &< \tilde{x}+ \tilde{w} / 2 < W - d_\text{min} \\
    d_\text{min} < \tilde{y} - \tilde{h} / 2 &< \tilde{y}+ \tilde{h} / 2 < H - d_\text{min}.
\end{align*}

\paragraph{Tracker measurements}
\label{subsubsec:valid_tracker}
At each iteration of the UKF, anticipated measurements are predicted
for a range of perturbations about the current state estimate. From
this set of possible measurements, a mean $\mathbf{\hat{z}}$ and
covariance $\mathbf{\hat{S}}$ representing the distribution of
observations is calculated. The Mahalanobis distance
(Eqn.~\ref{eqn:mahalanobis}) is compared against a threshold to
evaluate the likelihood of new measurements. If a mismatch between the
state estimate of the back-end and the new measurement is detected,
this could indicate drift in the tracker or some other failure, and
detection is
triggered. 

Additionally, to ensure the object is fully in view, $z$-tests using
the diagonal values of $\mathbf{\hat{S}}$ are performed at each
edge. The results are compared with a threshold.

The probabilistic thresholds in this section can be chosen from the
quantiles of the chi-squared distribution and normal distribution,
respectively. When the test fails, the system switches into detection
mode to either confirm it should stop tracking the object, or to
correct its tracked bounding box.

\section{Experimental Results}

\subsection{Implementation choices}
For our experiments we implement MSL-RAPTOR with
YOLOv3~\cite{redmon2018yolov3} from
Ultralytics~\cite{glenn_jocher_2019_2672652} for our front-end's
object detector. We chose YOLOv3 due to its ubiquity, speed, and
demonstrated ability to detect a range of objects. To support the
objects used in our experiments, we retrain it on the COCO dataset
mixed with a custom dataset containing images of a flying quadrotor,
which leads to 81 supported object categories.  For our visual
tracker, we rely on the implementation of
SiamMask~\cite{wang2019siammask} provided by the authors, which is
object-agnostic and therefore does not require retraining. We chose
SiamMask for its speed, ability to produce angled bounding boxes, and
impressive tracking performance. We use the provided weights trained
on the VOT dataset~\cite{VOT_TPAMI}.

\subsection{Evaluation on the NOCS-REAL275 dataset}

%
%
We begin by evaluating MSL-RAPTOR on the NOCS-REAL275
dataset~\cite{wang2019normalized} to answer a fundamental question
about the method: can the algorithm track pose accurately for a range
of objects? In all, the dataset contains seven training videos and six
testing videos from a camera moving in different scenes. The scenes
contain objects with class labels in \{\texttt{bowl}, \texttt{mug},
\texttt{can}, \texttt{bottle}, \texttt{laptop}, and
\texttt{camera}\}. We tune our class-specific UKF parameters using the
training set and present the results evaluated on the testing
set. Similarly to the baseline methods, we do not penalize orientation
errors about axes of symmetry for the \texttt{can}, \texttt{bowl}, and
\texttt{bottle} classes.

\paragraph{Implementation details}
Our re-detection and verification mechanisms presented in
\ref{subsec:update_tracked_objects} serve to make the tracking robust
to occlusions and to handle objects exiting and entering the
frame. However, sequences of the dataset are relatively short and
mostly maintain all objects in the field of view. Moreover, the
datasets used to train our detector did not contain instances of all
the classes appearing in the scenes, and the training set from
NOCS-REAL275 would not be general enough to train the detector to
properly detect new instances of objects in the testing set. For these
reasons, we disable the probabilistic detection and verification
mechanisms for evaluation on this dataset and rely only on visual
tracking in the front-end. This is a conservative approach as
violations of these assumptions will reduce our performance.

To replace MSL-RAPTOR's filter initialization, which is based on the
disabled detection, the same method is used as the
baselines~\cite{wang20196sixpack, Issac_2016, Wuthrich_2013}. For a
given object $O$, random uniform translation noise of up to 4cm is
added to the ground truth initial pose $\mathbf{T}^0$. From this
initialized pose, the first bounding box is calculated using the UKF's
measurement prediction function
$\mathbb{\pi}_\mathbf{K}(\mathbf{B}, \mathbf{T}^0)$. We add uniformly
sampled noise of up to 5\% of the width and height of the box to its
row and width, and its column and height respectively. With this first
bounding box, the visual tracker will generate future measurements.

For objects in the dataset with shapes not well approximated by a 3D
bounding box (bowl, laptop, mug), we add additional vertices to
restrict the volume. For example, 12 points are used instead of 8 for
the laptop, resulting in an ``L" shape when viewed from the side. For
mugs and bowls, extra points allow a tighter fit around curved
rims. Using these refined volumes reduces the mismatch between
predicted and observed measurements in the UKF, which can lead to
incorrect adjustments of the pose estimate. This effect is greatest
when objects are close to the camera, as in the NOCS dataset.

Finally, as the baseline methods are evaluated using desktop GPUs, for
this set of experiments only we use a desktop computer with an Nvidia
RTX 2060 GPU.

\paragraph{Baselines}
We compare our results with those reported by Wang \etal over the same
dataset~\cite{wang20196sixpack} for four methods, \textbf{all of which
  use RGB-D input}:
\begin{itemize}
\item NOCS~\cite{wang2019normalized}, a recent 6D pose estimation
  solution relying on pixel correspondences to a canonical
  representation.
\item ICP~\cite{zhou2018open3d}, the Open3D implementation of the
  iterative closest point algorithm.
\item KeypointNet~\cite{suwajanakorn2018discovery}, a category-based
  3D keypoint generator.
\item 6-PACK~\cite{wang20196sixpack}, a state-of-the-art method using
  learned 3D keypoints with an anchoring mechanism incorporating
  temporal information for improved tracking.
\end{itemize}

\paragraph{Results}
\begin{table}[t]
\centering
\small
\caption{Results on NOCS-REAL275. Translations are in cm and rotations are in degrees. }
\label{tab:nocs}
\scalebox{0.9}{
\begin{tabular}{|c|c||c c|c c|c c|c c|c c|c c||c c|}
\hline
\multirow{2}{*}{Method} & \multirow{2}{*}{Modality} & \multicolumn{2}{c|}{Bottle} & \multicolumn{2}{c|}{Bowl} & \multicolumn{2}{c|}{Camera} & \multicolumn{2}{c|}{Can} & \multicolumn{2}{c|}{Laptop} & \multicolumn{2}{c|}{Mug} & \multicolumn{2}{c|}{Overall} \\ \cline{3-16} 
                        &                           & $R_{err}$    & $t_{err}$    & $R_{err}$   & $t_{err}$   & $R_{err}$    & $t_{err}$    & $R_{err}$   & $t_{err}$  & $R_{err}$    & $t_{err}$    & $R_{err}$   & $t_{err}$  & $R_{err}$     & $t_{err}$    \\ \hline \hline
NOCS~\cite{wang2019normalized}                    & RGB-D                     & 25.6         & 1.2          & 4.7         & 3.1         & 33.8         & 14.4         & 16.9        & 4.0        & 8.6          & 2.4          & 31.5        & 4.0        & 20.2          & 4.9          \\ \hline
ICP~\cite{zhou2018open3d}                     & RGB-D                     & 48.0         & 4.7          & 19.0        & 12.2        & 80.5         & 15.7         & 47.1        & 9.4        & 37.7         & 9.2          & 56.3        & 9.2        & 48.1          & 10.5         \\ \hline
Keypoint Net~\cite{suwajanakorn2018discovery}            & RGB-D                     & 28.5         & 8.2          & 9.8         & 8.5         & 45.2         & 9.5          & 28.8        & 13.1       & 6.5          & 4.4          & 61.2        & 6.7        & 30.0          & 8.4          \\ \hline
6-PACK~\cite{wang20196sixpack}                   & RGB-D                     & 15.6         & 1.7          & 5.2         & 5.6         & 35.7         & 4.0          & 13.9        & 4.8        & 4.7          & 2.5          & 21.3        & 2.3        & 16.0          & 3.5          \\ \hline \hline
Ours                    & \textbf{Mono.}        & 18.7         & 9.8          & 16.2        & 4.0         & 23.4         & 8.5          & 17.6        & 9.0        & 17.4         & 11.6         & 35.3        & 7.7        & 21.8          & 8.2          \\ \hline
\end{tabular}}
\end{table}  
\begin{figure}[tpb]%
\centering
\subfigure[][]{%
\includegraphics[width=0.32\linewidth]{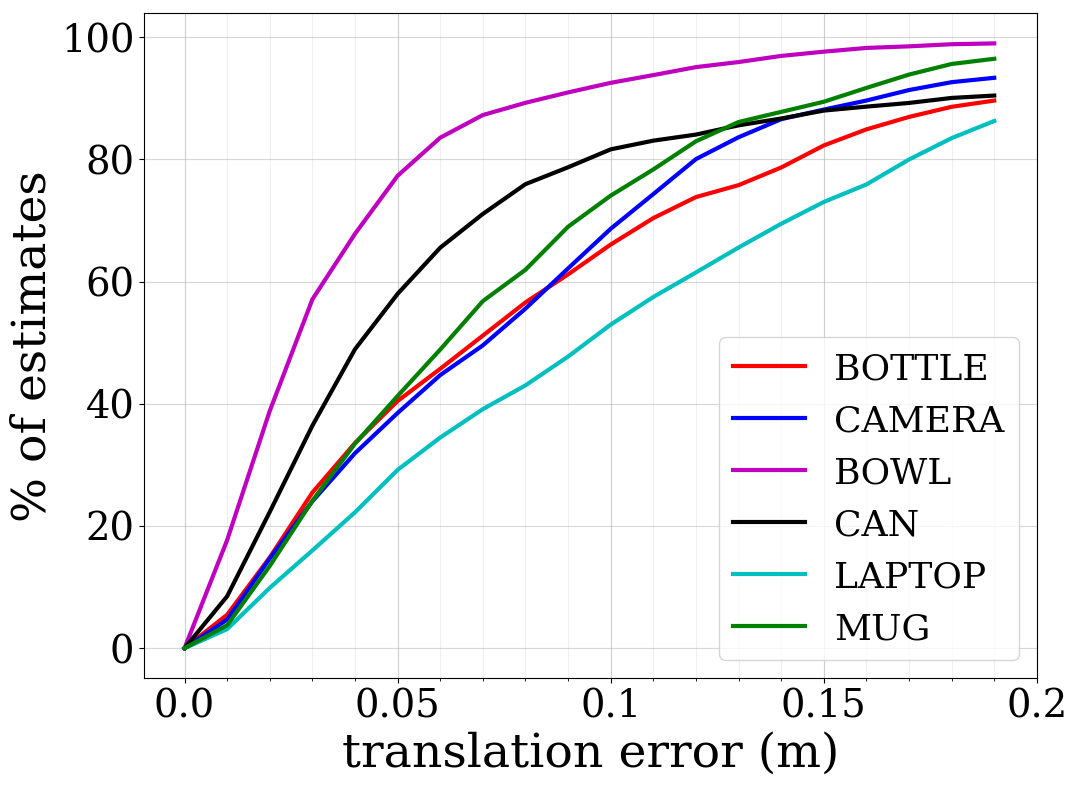}
\label{fig:nocs_s_curves-a}}
\subfigure[][]{%
\includegraphics[width=0.32\linewidth]{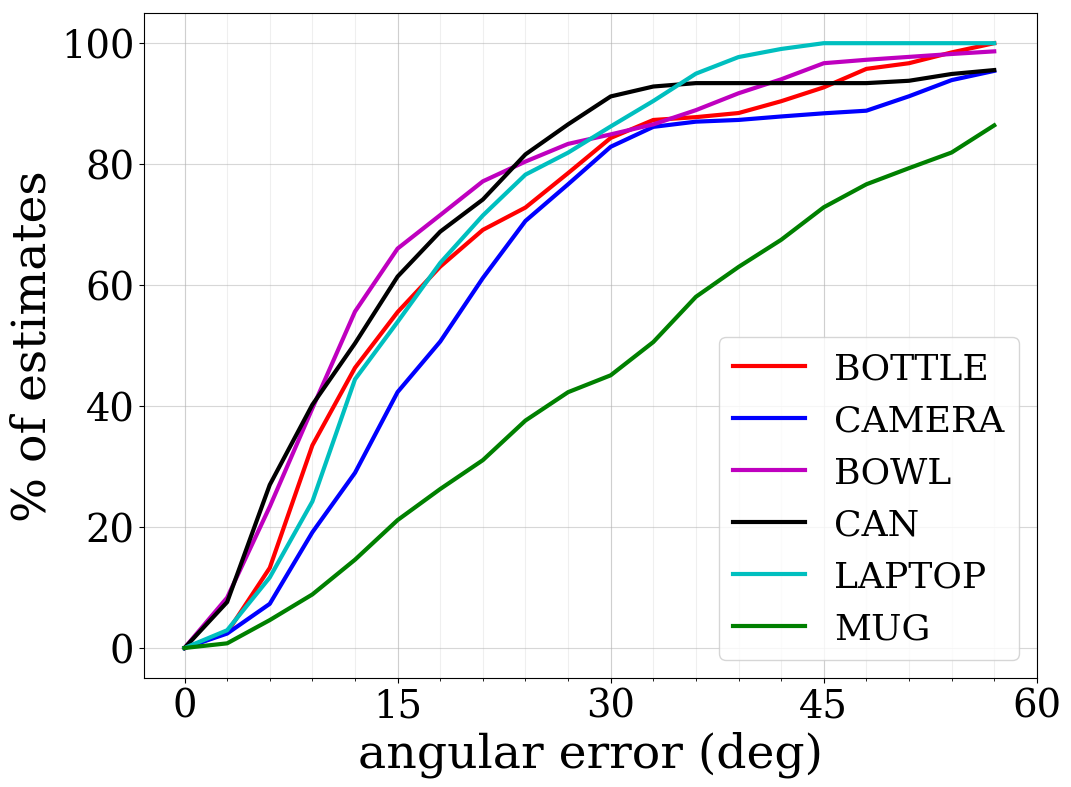}
\label{fig:nocs_s_curves-b}} 
\subfigure[][]{%
\includegraphics[width=0.31\linewidth]{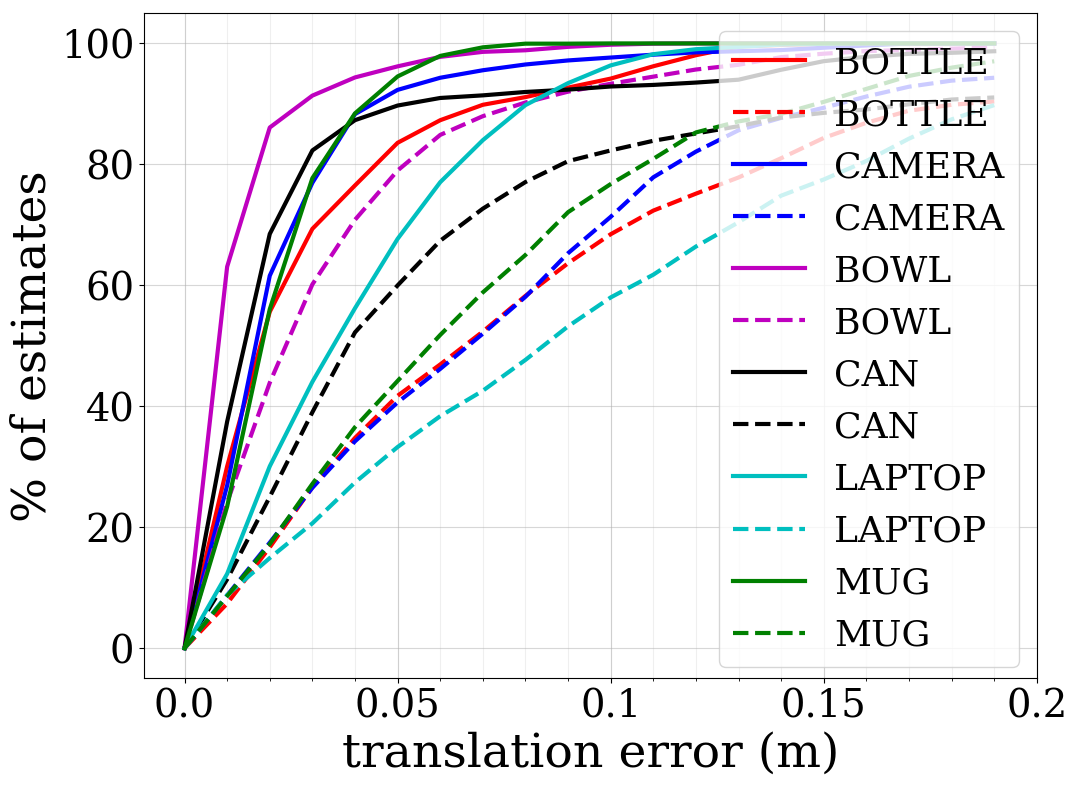}
\label{fig:nocs_s_curves-c}}
\caption{Empirical cumulative distribution function on errors:
\subref{fig:nocs_s_curves-a} translation only;
\subref{fig:nocs_s_curves-b} rotation only;
\subref{fig:nocs_s_curves-c} translation in-plane error (solid line) and depth error (dashed line)}%
\label{fig:nocs_s_curves} 
\end{figure}
The average rotation and translation errors over each class are
presented in Tab.~\ref{tab:nocs} and the accuracy visualized in
Fig.~\ref{fig:nocs_s_curves}. Despite relying only on monocular
images, MSL-RAPTOR achieves comparable performance to state-of-the-art
methods that use RGB-D. As is expected for a monocular camera, depth
estimates are less accurate than translations perpendicular to the
agent's camera axis (i.e. parallel to the image plane). This can be
seen in Fig.~\ref{fig:nocs_s_curves-c} where these are broken out into
separate errors. This is a problem that comes with the use of
monocular images rather than RGB-D data.

\subsection{Baseline evaluation in an aerial robotics scenario}
\begin{figure}[]
\centering
\includegraphics[clip, trim=5.75cm 1.5cm 4.5cm 1.75cm, width=0.35\linewidth]{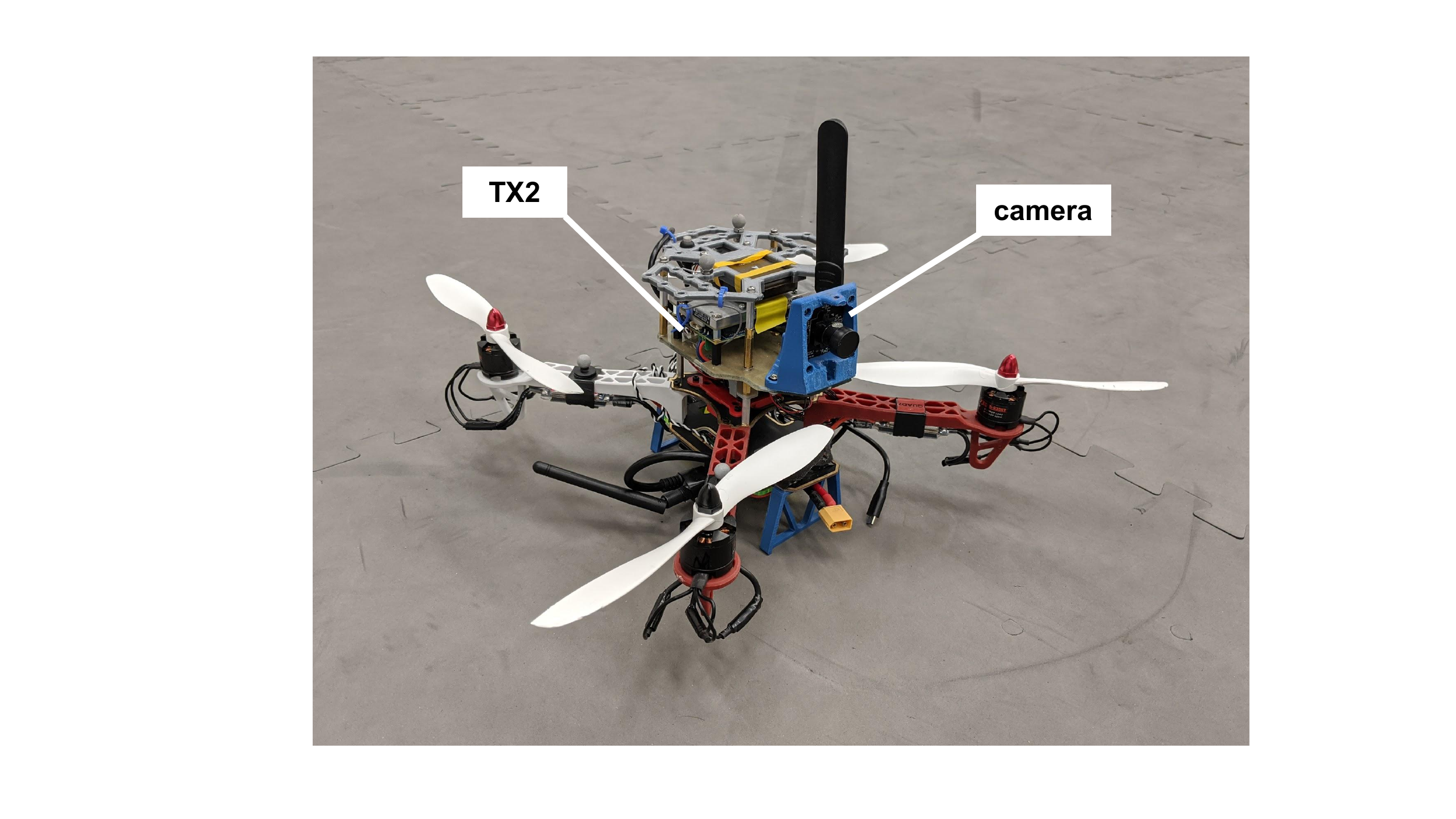} 
\caption{Quadrotor used for tracking objects in robotic perception scenarios.}
\label{fig:tx2_drone}
\end{figure}

To demonstrate that MSL-RAPTOR provides robust onboard perception, we
ran the algorithm onboard a drone with an Nvidia Jetson TX2 to track
another drone. The ego quadrotor uses the DJI F330 frame and is
equipped with off-the-shelf and custom made components as shown in
Fig.~\ref{fig:tx2_drone}. The quadrotor uses an RGB camera (640
$\times$ 480 resolution). Experiments were designed so that the ego
drone moves about the experimental space causing the image background
to change substantially, a challenge which demonstrates the robustness
of the system. The flights varied from slow to aggressive, with the
tracked object occasionally exiting and reentering the field of view
or being occluded by other objects. As would be expected in collision
avoidance situations, the distances between the ego and tracked drone
routinely reached 8 meters (a substantially larger separation distance
than what is seen in most existing pose estimation datasets). This
provides important validation of the use of this method for planning
and decision making onboard agile robots. From our collected data, we
use 5500 images over several runs to form a training set for the
detector and baseline, and use 5000 different images to form a testing
set to evaluate pose predictions.

\paragraph{Implementation details}
We do not assume knowledge about the control input of the tracked
drones, and a simple constant velocity dynamics model is used. Since
we are looking at tracking the other drone at up to 10 meters distance
from the ego drone, we treat the tracked drone as symmetric about the
vertical axis considering the yaw angle does not have an appreciable
effect on our measurements, and does not provide useful information
about its behavior. For a fair comparison, the same assumption is made
for the baseline method. With a higher resolution camera or for
tracking at closer ranges this constraint could be removed.

\paragraph{Baseline}
While most state-of-the-art methods struggle to achieve real-time
speeds on desktop GPUs, SSP~\cite{tekin2018ssp} reports inference-only
speeds of 50 Hz. This, in addition to its impressive accuracy, makes
it the only candidate likely to be competitive when running
onboard. SSP takes in an image and outputs the estimated locations of
the projected 3D bounding box of the object on the image plane. Its
network is based on YOLO's architecture and is modified to produce the
3D bounding box projections instead of a 2D bounding box. The relative
rigid body transformation is then calculated from the projected vertex
locations and the 3D bounding box geometry using a PnP algorithm.

SSP requires the full object pose for each image when
training. Outside of laboratory environments with specialized
equipment, this is more challenging to acquire than the 2D bounding
boxes used by MSL-RAPTOR. Moreover, SSP is presented as an
instance-based method, meaning it must be trained on specific objects
for which it will infer pose. For this reason, we did not evaluate it
on the NOCS dataset that contains unseen objects instances in the
testing set.

We train SSP using a desktop GPU, and present results for both SSP and
MSL-RAPTOR evaluated on an Nvidia Jetson TX2.
\paragraph{Results}
\begin{figure}[tpb]%
\centering
\subfigure[][]{%
\includegraphics[width=0.32\linewidth]{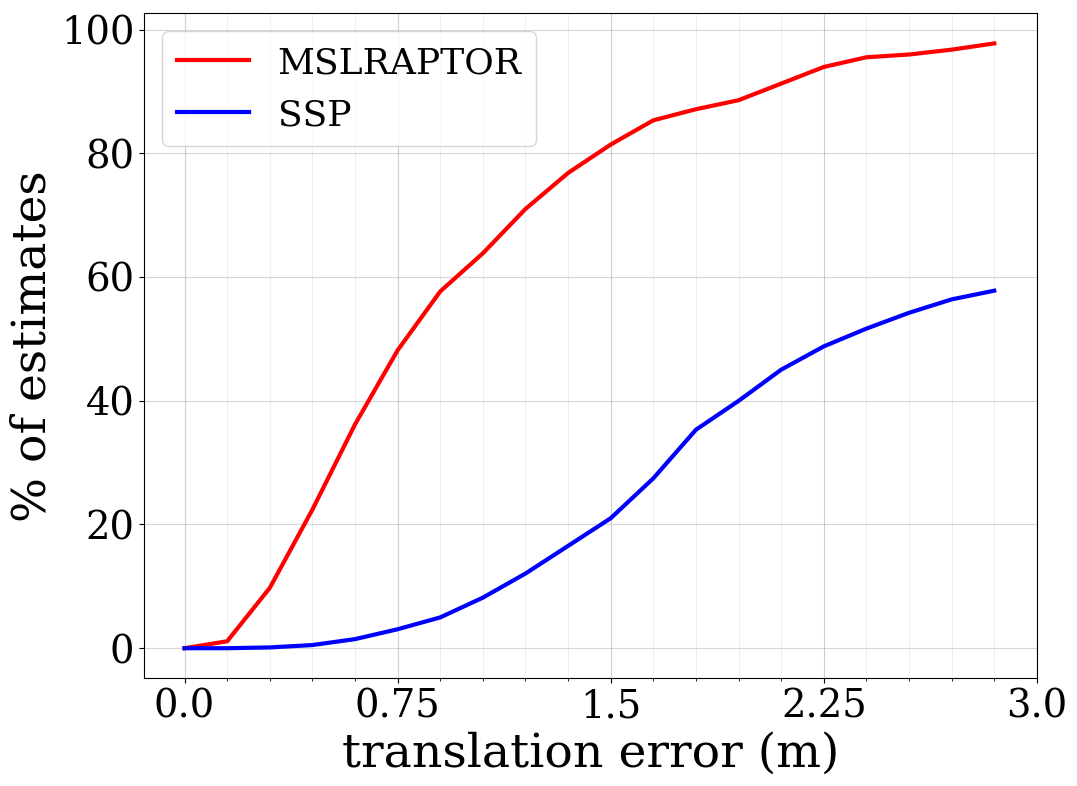}
\label{fig:ssp_s_curves-a}}
\subfigure[][]{%
\includegraphics[width=0.32\linewidth]{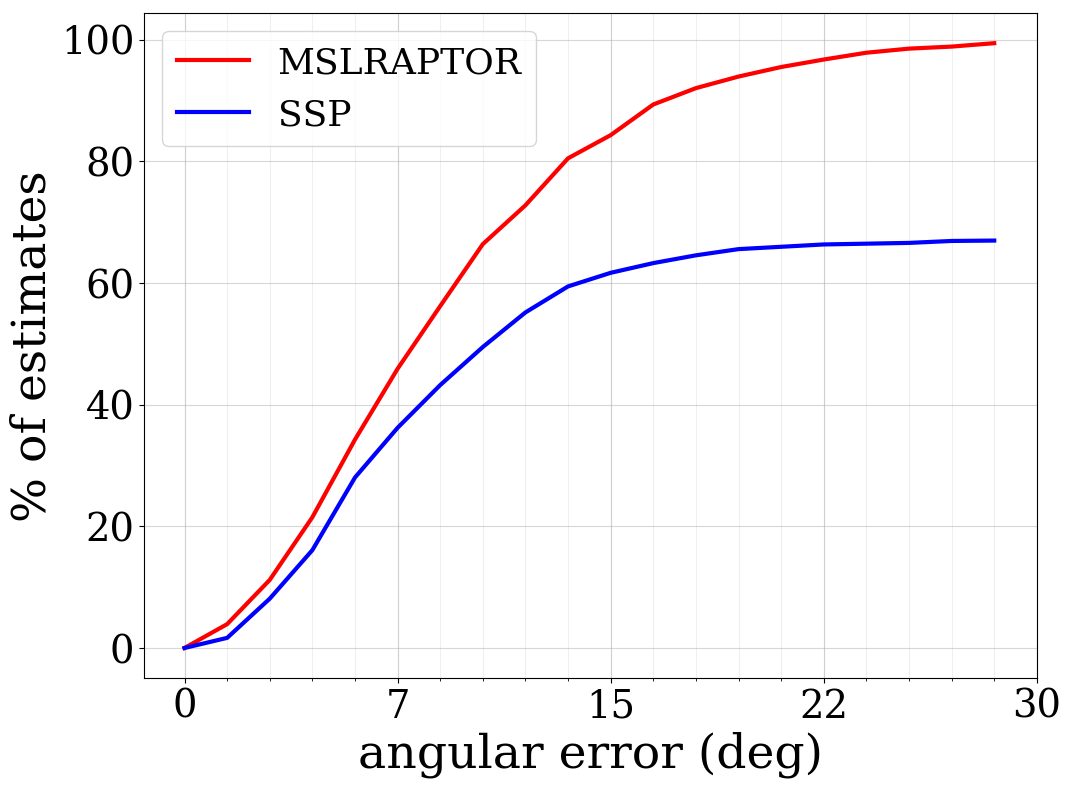}
\label{fig:ssp_s_curves-b}} 
\subfigure[][]{%
\includegraphics[width=0.32\linewidth]{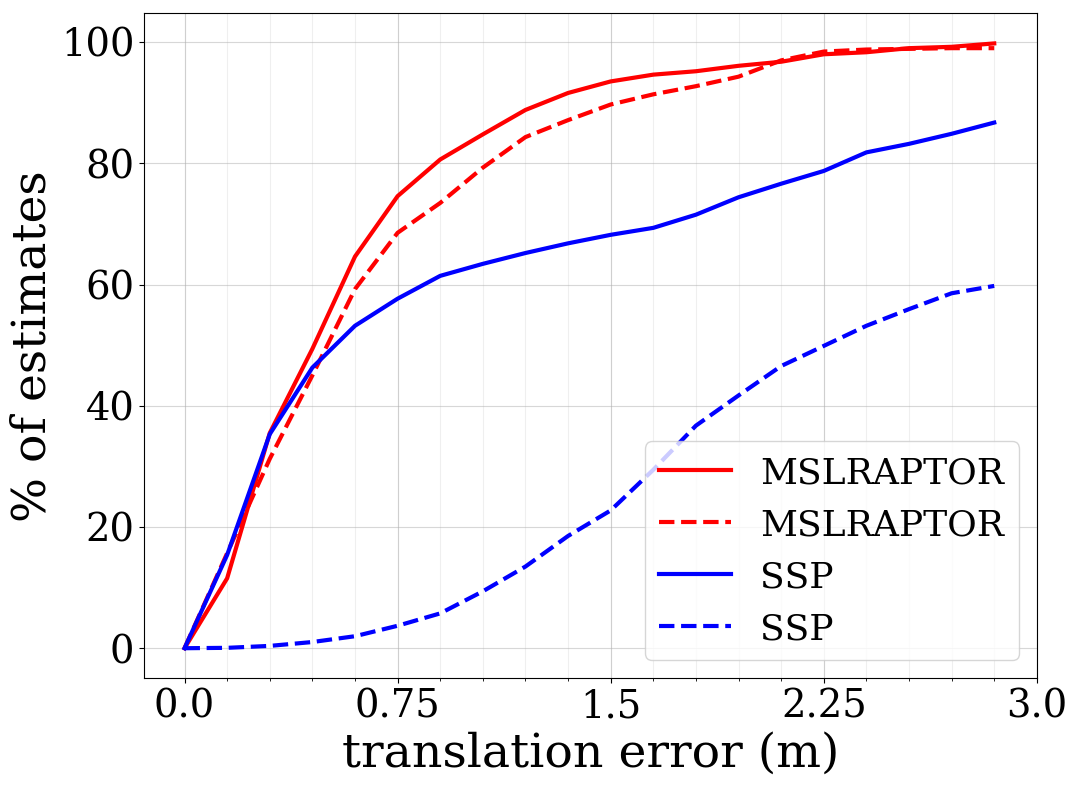}
\label{fig:ssp_s_curves-c}}
\caption{Empirical cumulative distribution function on errors:
\subref{fig:ssp_s_curves-a} translation only;
\subref{fig:ssp_s_curves-b} rotation only;
\subref{fig:ssp_s_curves-c} translation in-plane error (solid line) and depth error (dashed line)}%
\label{fig:ssp_s_curves} 
\end{figure}
Fig.~\ref{fig:ssp_s_curves} shows the better accuracy of MSL-RAPTOR
compared to SSP. The median translation and rotation errors are 0.82 m
and 8.59 degrees for MSL-RAPTOR compared to 2.45 m and 11.26 degrees
for SSP (66\% less translation error and 23\% less rotation
error). When tracking a drone, MSL-RAPTOR runs 3X faster than SSP on a
TX2, with an average pose prediction speed of 0.115 s (8.7 Hz)
compared to 0.335 s (2.98 Hz) for SSP.

\begin{figure}[tpb]%
\centering
\subfigure[][]{%
\includegraphics[clip, trim=11.25cm 7.5cm 7.5cm 7.5cm, width=0.30\linewidth]{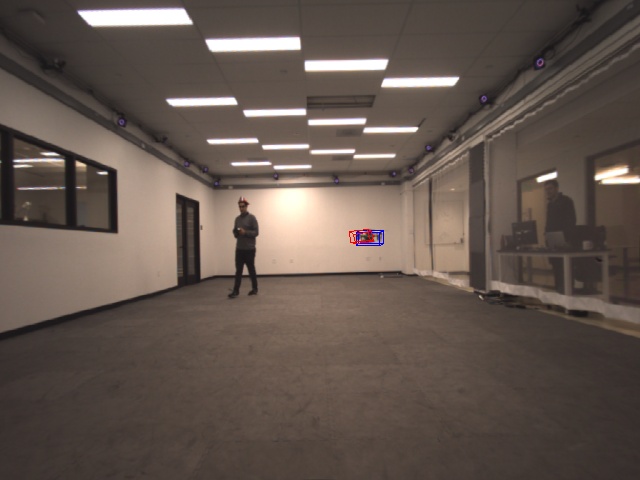}
\label{fig:ssp-fail-a}}
\hspace{20pt}%
\subfigure[][]{%
\includegraphics[clip, trim=11.25cm 7.5cm 7.5cm 7.5cm, width=0.30\linewidth]{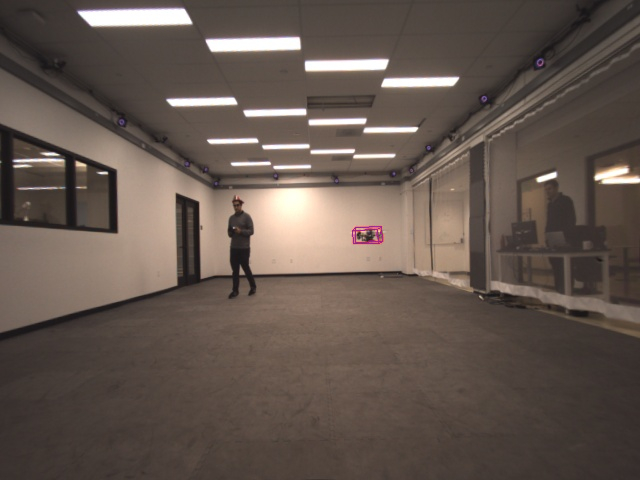}
\label{fig:ssp-fail-b-b}} 
\caption{Left: SSP's prediction for the projected 3D bounding box
  (red), and the ground truth box (blue). Note that due to the fast
  motion of the drone, the blurry image results in a distorted
  prediction. Right: MSL-RAPTOR's prediction for the same frame
  (purple).}%
\label{fig:ssp_fail} 
\end{figure}

SSP directly outputs the locations of the 3D bounding box projections
on the image, so when the drone is far away or moving quickly (causing
blur), these points can be warped and scaled
(Fig.~\ref{fig:ssp_fail}). Regardless, the PnP algorithm will find a
rigid body transformation that best matches these projections to the
actual 3D bounding box, leading to significant depth errors as seen in
Fig.~\ref{fig:ssp_s_curves-c}. Since the in-plane translation is
affected by the average of these points, and the rotation by the
relative positioning, the difference in performance is less stark for
the rotation and in-plane translation errors. As SSP does not
incorporate temporal information it cannot identify nor correct for
these situations which lead to estimates with large errors, resulting
in the apparent asymptotes in Fig.~\ref{fig:ssp_s_curves}.

\section{Conclusions} 

\label{sec:conclusion}
In this paper, we propose MSL-RAPTOR, an efficient monocular relative
pose tracking algorithm that combines the image processing
capabilities of modern neural networks with the robustness advantages
of classical probabilistic filters. The front-end extracts class
labels and angled bounding boxes, which are used as observations by
the back-end's UKF. The back-end estimates the relative pose and
returns measurement uncertainty parameters to the front-end for
methodically determining when to trigger the slower, but more
accurate, re-detection. When not detecting, the faster visual tracking
method is used. The algorithm runs 3 times faster than the quickest
state-of-the-art method, while still out-performing it in terms of
precision in onboard robotic perception scenarios. Furthermore,
despite not using depth measurements, comparable performance to
state-of-the-art results is achieved on the NOCS-REAL275 dataset.


\subsubsection*{Acknowledgements} 
This research was supported in part by ONR grant number N00014-18-1-2830, NSF NRI grant 1830402, the Stanford Ford Alliance program, the Mitacs Globalink research award IT15240, and the NSERC Discovery Grant 2019-05165. We are grateful for this support.

\end{document}